\pdfoutput=1

\documentclass[11pt]{article}

\usepackage[]{EMNLP2023}

\usepackage{times}
\usepackage{latexsym}

\usepackage[T1]{fontenc}

\usepackage[utf8]{inputenc}

\usepackage{microtype}

\usepackage{inconsolata}
\usepackage{float}
\usepackage{amsfonts}
\usepackage{pdfpages}

\usepackage{color}
\usepackage{algorithm}
\usepackage{algorithmic}
\usepackage{amsmath}
\usepackage{amsthm}
\usepackage{enumitem}
\usepackage{bm}
\usepackage{subfigure} 
\usepackage{nicefrac}
\usepackage{tabularx}
\usepackage{url}
\usepackage{booktabs}
\usepackage{multirow}
\usepackage{graphicx}
\usepackage{amssymb}
\usepackage{bbding}

\graphicspath{ {./images/} }

\newcolumntype{L}[1]{>{\raggedright\let\newline\\\arraybackslash\hspace{0pt}}m{#1}}
\newcolumntype{C}[1]{>{\centering\let\newline  \\\arraybackslash\hspace{0pt}}m{#1}}
\newcolumntype{R}[1]{>{\raggedleft\let\newline \\\arraybackslash\hspace{0pt}}m{#1}}

\title{Heterogeneous Graph Neural Networks with \\ Loss-decrease-aware Curriculum Learning }

\author{Yili Wang$^a$ \\
$^a$  Department of Computer Science and Engineering, Central South University , \\
\texttt{\small wangyili@csu.edu.cn, }
}

\begin{document}
\maketitle
\begin{abstract}
In recent years, heterogeneous graph neural networks (HGNNs) have achieved excellent performance
in handling heterogeneous information networks (HINs).
Curriculum learning is a machine learning strategy where training examples are presented to a model in a structured order, starting with easy examples and gradually increasing difficulty, aiming to improve learning efficiency and generalization.
To better exploit the rich information in HINs, previous methods have started to explore the use of curriculum learning strategy to train HGNNs.
Specifically, these works utilize the absolute value of the loss at each training epoch to evaluate the learning difficulty of each training sample. 
However, the relative loss, rather than the absolute value of loss, reveals the learning difficulty.
Therefore, we propose a novel loss-decrease-aware training schedule (LDTS).
LDTS uses the trend of loss decrease between each training epoch to better evaluating the difficulty of training samples, thereby enhancing the curriculum learning of HGNNs for downstream tasks.
Additionally, we propose a sampling strategy to alleviate training imbalance issues.
Our method further demonstrate the efficacy of curriculum learning in enhancing HGNNs capabilities.
We call our method \textbf{L}oss-\textbf{d}ecrease-aware \textbf{H}eterogeneous \textbf{G}raph \textbf{N}eural \textbf{N}etworks (LDHGNN).
The code is public at \url{https://github.com/wangyili00/LDHGNN}. 

%
\end{abstract}

\section{Introduction}

Heterogeneous information networks (HINs) \cite{HINs} have been extensively utilized across various real-world applications. 
Featuring diverse types of relations linking different node types, HINs offer an ideal framework for representing complex datasets from domains such as social networks \cite{socialnet1,socialnet2}, e-commerce \cite{e-commerce}, and quantum chemistry \cite{comprehensive}.
To better exploit the rich information in this complex structure, Heterogeneous Graph Neural Networks (HGNNs) \cite{RGCN, HGT, mao2023hinormer,SeHGNN} have emerged and achieved excellent performance.

Despite the success of HGNNs, current research primarily focuses on model design, with limited attention paid to model training, which Hinder the further improvement of HGNN performance.
To address this challenge, CLGNN \cite{CLGNN} has been proposed based on curriculum learning.
Specifically,  CLGNN designing a evaluator to evaluate the learning difficulty of each training sample through the absolute value of the loss.
Then they employ a linear growth strategy to train the model in a progressive manner that increases difficulty step by step.
However, the CLGNN still has the following deficiencies. 
\textbf{On one hand,} the relative loss between each training epoch, rather than the absolute value of loss, reveals the learning difficulty of sample.
In simple terms, easily learnable samples typically exhibit a significant decrease in loss after a single model update.
Therefore, using the absolute value of loss as a difficulty evaluator in CLGNN may misguide curriculum learning, leading to sub-optimal performance.
\textbf{On the other hand,} CLGNN may overfit to easily learnable samples and underfit to difficult samples, since it tends to completely ignore difficult samples for the majority of the training phase.
As a result, this also lead to suboptimal performance.

In view of these limitation, we propose Loss-decrease-aware Heterogeneous Graph Neural Networks (LDHGNN) to further enhance the performance of HGNNs.
\textbf{To address the first issue}, we propose loss-decrease-aware training schedule (LDTS).
Specifically, LDTS uses the trend of loss decrease between each training epoch to better evaluating the difficulty of training samples, thereby enhancing the curriculum learning of HGNNs for downstream tasks.
\textbf{To address the second issue}, we propose a novel sampling strategy.
This strategy convert loss-decrease into probability, and then samples based on these probabilities.
In this way, we can mitigate overfitting or underfitting caused by training imbalance.
Experiment demonstrate the effectiveness of our method.

\section{Related Work}
\subsection{Heterogeneous Graph Neural Networks}
Graph Neural Networks (GNNs) are a type of neural network specifically designed to operate on graph-structured data. 
As one of the most representative papers in the field of GNNs, the Graph Convolution Network (GCN) \cite{kipf2016semi} has garnered significant attention in recent years due to its effectiveness.
GCN employ the message passing mechanism to learn local topological structure features of graphs,
thereby obtaining node representations.

To generalize the success of homogeneous GNNs \cite{kipf2016semi,GAT,EDGE,SAGE}, 
recently, there has been a growing number of proposals for Heterogeneous Graph Neural Networks (HGNNs). 
These methods enhance the performance of HGNNs by continuously improving the design of aggregation operations RGCN \cite{RGCN, HGT, mao2023hinormer,SeHGNN} or by designing specialized message aggregation path \cite{HAN,GTN,DiffMG,wang2024graph}.

Furthermore, to efficiently process large-scale heterogeneous graphs, some researches \cite{SeHGNN, li2023long, hu2023efficient} have attempted to decouple the aggregation operations from the model training stage to achieve faster training speeds.
Typically, Random Projection Heterogeneous Graph Neural Network (RpHGNN) \cite{hu2023efficient} merges efficiency with low information loss by employing propagate-then-update iterations. It features a Random Projection Squashing step for linear complexity growth and a Relation-wise Neighbor Collection component with an Even-odd Propagation Scheme for finer neighbor information gathering.

Current HGNN method primarily focuses on model design, with limited attention paid to model training, which Hinder the further improvement of HGNN performance. 
To address this challenge, CLGNN \cite{CLGNN} has been proposed based on curriculum learning.
CLGNN designing a evaluator to evaluate the learning difficulty of each training sample through the absolute value of the loss. 
Then they employ a linear growth strategy to train the model in a progressive manner that increases difficulty step by step.


\subsection{Curriculum Learning}
Curriculum learning (CL) \cite{curriculum,curriculum2,curriculum3} is a widely recognized approach inspired by educational psychology, where the learning process gradually increases in complexity or difficulty of tasks presented to the model. 
This approach mirrors human learning by introducing training data from simpler to more complex samples. 
A typical CL framework includes a difficulty measurer and a training schedule. 
The CL strategy, serving as a user-friendly plugin, has proven effective in enhancing the generalization capability and convergence speed of various models across a wide range of scenarios such as computer vision and natural language processing, among others. 
However, accurately assessing the difficulty or quality of nodes within a graph with specific topology remains an unresolved challenge.

\section{Proposed Methods}
In this section, we will introduce the detail of our method.
We primarily introduce the proposed loss-decrease-aware training schedule and sampling strategy.
\subsection{Loss-decrease-aware Heterogeneous Graph Neural Networks}
\textbf{Loss-decrease-aware training schedule.} 
The topology of a graph is intricate and complex, making it difficult to ensure that various models effectively assess the difficulty and quality of nodes based solely on topological features.
Previous work simply used the loss of each node as a metric to evaluate difficulty.
However, the relative loss between each training epoch, rather than the absolute value of loss, reveals the learning difficulty of sample.
Therefore, we propose loss-decrease-aware training schedule (LDTS) to more effectively evaluate difficulty of each sample. 
Specifically, in each training epoch, we calculate the loss for all samples and compare it with the training loss from the previous iteration.
Samples with a significant decrease in loss compared to the previous iteration are considered easy samples, while those with less decrease or an increase in loss are considered difficult samples.
Then, we use softmax operation to convert loss decrease into probability, which implies that easy samples will have a higher probability of being selected.
Finally, We will sample nodes based on their probabilities, and the number of sampled nodes will increase gradually with the epoch.
Only the nodes that are sampled will have their losses backpropagated to train the model.
The algorithm is outlined in Algorithm~\ref{alg:alg}:


  \begin{algorithm}[ht]
    \caption{Loss-decrease-aware training schedule (LDTS)}
    \label{alg:alg}
    \begin{algorithmic}[1]
    \STATE \textbf{Input}: HGNN model $M()$; Node feature $X$; Label $Y$.\\
    \STATE \textbf{Initialization}: $preloss = [0,...0]$\\
    \STATE \textbf{Model training}: \\
    \WHILE{model not converged}
    \STATE \textbf{Model forward calculation}: \\
    $H = M(X)$
    \STATE \textbf{Calculate loss for each node}: \\
     $loss = loss\_func(H, Y, 'none')$
     \STATE \textbf{Calculate loss decrease}: \\
     $D = preloss - loss$ \\
     $preloss = loss $
     \STATE \textbf{Convert loss decrease into probability}: \\
     $P = softmax(D)$ 
     \STATE \textbf{Strategize Training Schedule}: \\
     $size = training\_schedule()$ \\
     $num\_large= len(loss) * size$\\
     \STATE \textbf{Sample based on probability}: \\
      $indices = Sample(P, num\_large)$ \\
      $loss = loss[idx].mean()$
      \STATE  \textbf{Loss backward}: \\
     loss.backward()
    \ENDWHILE
      \STATE  \textbf{Return}: Trained HGNN Model\\
    \end{algorithmic}
  \end{algorithm}

A training schedule function maps each training epoch $t$ to a
scalar $\lambda_t \in (0, 1]$, indicating the proportion of the easiest training nodes utilized at the $t$-th epoch. Denoting $\lambda_0$ as the initial proportion of available easiest nodes and $T$ as the epoch when the training schedule function first reaches 1, three pacing functions are considered: linear, root, and geometric.The details of training schedule is defined as in Algorithm~\ref{alg:TS}:

  \begin{algorithm}[ht]
    \caption{Strategize Training Schedule}
    \label{alg:TS}
    \textbf{def}\quad training\_schedule(lam,t,T,scheduler):
    \begin{algorithmic}[1]
    \IF{$scheduler == 'linear'$} \RETURN {$min(1, lam + (1 - lam) * t / T)$} \ELSIF{$scheduler == 'root'$} \RETURN {$min(1, sqrt(lam^2 + (1 - lam^2) * t / T))$}\ELSIF{$scheduler == 'geom'$} 
    \RETURN {$min(1, 2 ^{(log2(lam) - log2(lam) * t / T)})$}\ENDIF
               
    \end{algorithmic}
  \end{algorithm}

Furthermore, training does not halt abruptly at $t=T$, as the backbone GNN may not have thoroughly explored the knowledge of recently introduced nodes at this juncture. Rather, for $t>T$, the entire training set is employed to train until the test accuracy on the validation set reaches convergence.

We can enhance the performance of any HGNN model by training it using the aforementioned algorithm. In this context, we used RpHGNN \cite{hu2023efficient} as the training model and use "linear" to strategize training schedule.

\section{Experiments}
In this section, we will thoroughly explore and scrutinize the performance of the proposed method. Through a variety of thorough experiments, we have demonstrated the effectiveness of our method by demonstrating increased accuracy in predictions.
\subsection{Experimental Setup}

\subsubsection{Dataset}
    We evaluate our method on the Open Graph Benchmark dataset, \textit{ogbn-mag}. 
    The \textit{ogbl-mag} dataset is a heterogeneous, directed graph which contains  four types of entities—papers, authors, institutions, and fields of study—as well as four types of directed relations connecting two types of entities—an author is “affiliated with” an institution, an author “writes” a paper, a paper “cites” a paper, and a paper “has a topic of” a field of study. Each paper is associated with a 128-dimensional word2vec feature vector, while no input node features are associated with other entity types.The objective is to identify the appropriate academic platform (either a conference or a journal) for each research paper, based on its content, citations, the names of the authors, and the institutions they are affiliated with. This task is particularly relevant since many submissions in the Microsoft Academic Graph (MAG) lack or have uncertain venue information due to the inherent inaccuracies in web-based data. The ogbn-mag dataset encompasses a vast array of 349 distinct venues, transforming this challenge into a multiclass classification task with 349 different classes.
    Statistics of the dataset are provided in Table~\ref{tab:dataset}.

\begin{table}[ht]
    \centering
    \caption{Dataset statistics}
    \small
    \begin{tabular}{c|ccc}
    \toprule
         Nodes type & Train & Validation & Test \\ \midrule
         Papers     & 629,571 & 64,879 & 41,939  \\ 
         Authors & 1,134,649   &  --  & -- \\
         Institutions & 8,740 &  -- & --\\
         Fields of study & 59,965 & --&--\\
    \bottomrule
    \end{tabular}
    \label{tab:dataset}
\end{table}

\subsubsection{Evalution Metric}
OGB provides standardized dataset splits and evaluators that allow for easy and reliable comparison of different models in a unified manner. For \textit{ogbl-mag} dataset, the evaluation metric is accuracy of multi-class classification. More specifically, the accuracy was determined by dividing the count of correct predictions across all 349 classes by the total number of predictions made for those classes.

\subsubsection{Settings}
The experiments were conducted on Ubuntu 20.04.4 LTS with NVIDIA RTX 3090 GPU and 128GB RAM.
To demonstrate the effectiveness of our approach.
We utilized the publicly available model codes provided by OGB (Open Graph Benchmark) along with their corresponding hyperparameters.


\begin{table}
    \centering
    \caption{Node Classification Performance of our method on dataset \textit{ogbn-mag}. }

    \small
    \setlength\tabcolsep{2 pt}
    \begin{tabular}{c|cc}
    \toprule
        Methods & Valid.  & Test  \\
        \midrule
         RpHGNN & 0.5973 ± 0.0008	& 0.5773 ± 0.0012 \\
         CLGNN  & 0.8021 ± 0.0020 & 0.7956 ± 0.0047 \\
         LDHGNN & \textbf{0.8836 ± 0.0028} & \textbf{0.8789 ± 0.0024} \\
         \bottomrule
    \end{tabular}
    \label{tab:accuracy improved}
\end{table}

\subsection{Accuracy Improvement}
As shown in Table~\ref{tab:accuracy improved}, we compared two models with our method.
RpHGNN \cite{hu2023efficient} represents the original model trained conventionally, and CLGNN \cite{CLGNN} represents training RpHGNN using curriculum learning, which is currently the state-of-the-art method on the ogbn-mag leaderboard.
Notably, our method outperforms the CLGNN method by nearly 8\%, demonstrating the effectiveness of our approach.

\section{Conclusion}
In this paper, we propose Loss-decrease-aware Heterogeneous Graph Neural Networks (LDHGNN) to further enhance the performance of HGNNs.
LDHGNN, based on curriculum learning, utilizes loss decrease as a measure of difficulty instead of the absolute value of loss, which can better assist curriculum learning in training the model.
Additionally, we employed a sampling strategy to mitigate the overfitting or underfitting issues caused by training sample imbalance in curriculum learning.
The experiments on real-world benchmark datasets \textit{ogbn-mag} validate the effectiveness of our method.
We aim to delve deeper into the notable results and validate its performance on various datasets through further experiments.

\bibliography{main}
\bibliographystyle{acl_natbib}

\end{document}